\documentclass[letterpaper, inpress]{jds} 

\jdsmonth{July}                 
\jdsyear{2020}                  
\jdsvolume{xx}                  
\jdsissue{xx}                   
\jdsdoi{xx.xxxx/xxxxxxxxx}      
\jdsreceived{April, 2020}       
\jdsaccepted{May, 2020}         

\usepackage{amsfonts,amsmath,amssymb,amsthm}
\usepackage{booktabs}
\usepackage{arydshln}
\usepackage{array,colortbl,multirow,multicol,booktabs,ctable}
\usepackage{lipsum}
\title[Abstractive Summarization With Semantic Graph]{Augmented Abstractive Summarization With Document-Level Semantic Graph}

\author[1]{Qiwei Bi}
\author[2]{Haoyuan Li}
\author[3]{Kun Lu}
\author[4,1]{Hanfang Yang\thanks{Corresponding author. Email: hyang@ruc.edu.cn}}

\affil[1]{School of Statistics, Renmin University of China, Beijing, China}
\affil[2]{T.H. Chan School of Public Health, Harvard University, Boston, MA, USA}
\affil[3]{Department of ORFE, Princeton University, Princeton, New Jersey, USA}
\affil[4]{Center for Applied Statistics, School of Statistics, Renmin University of China, Beijing}

\begin{document}

\maketitle

    \begin{abstract}
        Previous abstractive methods apply sequence-to-sequence structures to generate summary without a module to assist the system to detect vital mentions and relationships within a document. To address this problem, we utilize semantic graph to boost the generation performance. Firstly, we extract important entities from each document and then establish a graph inspired by the idea of distant supervision \citep{mintz-etal-2009-distant}. Then, we combine a Bi-LSTM with a graph encoder to obtain the representation of each graph node. A novel neural decoder is presented to leverage the information of such entity graphs. Automatic and human evaluations show the effectiveness of our technique.

    \end{abstract}
    
        \section{Introduction}
    Text summarization systems aim to condense a piece of text into a shorter summary, while preserving the key information and the meaning of the content. After showing superior performance in machine translation \cite{bahdanau2014neural}, the sequence-to-sequence models have also made encouraging progress in abstractive summarization \cite{rush-etal-2015-neural,nallapati-etal-2016-abstractive}.
    
    However, simply adopting this structure in the summarization task is frequently found to bring undesirable behavior such as inaccurate factual details, repetitions, and lack of coherence \cite{see-etal-2017-get,paulus2018a}. This is mainly because the length of the input text for summarization is in general much longer than that of the machine translation task, while the generated summary is much shorter than the original document. Without good guidance, the model will have difficulty in choosing which aspects of content to focus on during generation. ~\citet{fan2018robust} suggests that existing methods have difficulty in identifying salient entities and related events. In order to alleviate this problem, ~\citep{sharma-etal-2019-entity,gehrmann-etal-2018-bottom} propose to follow a two-step schema by first selecting important phrases and then paraphrasing them. However, the corpus does not provide the related `importance label' which leads to the loss of information in the first stage.
    
 Structured data is helpful to tackle this issue. Abstract Meaning Representation (AMR), proposed by ~\citet{banarescu-etal-2013-abstract}, is a semantic graph representation of a sentence that can grasp the vital entities and relations. ~\citet{damonte-cohen-2019-structural} show the benefits of graph encoders in AMR-to-text generation, which aims to generate sentences from AMR graphs. Semantic graphs also prove to be very useful in generating longer texts expressing complex ideas. ~\citet{logan-etal-2019-baracks} introduce a language model conditioned on an external knowledge graph relevant to the context. ~\citet{koncel2019text} propose an end-to-end trainable system for graph-to-text generation. However, due to the lack of golden annotations, introducing graph-related methods to summarization is difficult.
    
 Intuitively, provided with a cheat sheet about the essential entities and their relationships, it would be much easier to generate an accurate and coherent summary. Inspired by this, in this paper, we use information extraction methods to extract entities, coreferences, and relations for each document. To extract entities and coreferences, we simply adopt the industrial-strength tool Spacy \citep{spacy2}. There are many state-of-the-art relation extraction models \citep{luan-etal-2018-multi,zhang-etal-2018-graph,trisedya-etal-2019-neural}, but these methods are all limited to a specific domain. If we want to apply these models in summarization, we have to collect related annotations and retrain them on new data, e.g. CNN/Daily mail (CNNDM) and New York Times (NYT). Therefore, we adopt a semi-supervised method to address this problem. Inspired by distant-supervision \citep{mintz-etal-2009-distant}, two types of relationships are considered in this paper: (1) two mentions from different coreference clusters are deemed to have relations if they appear in the same sentence; (2) two entities that participate in a known knowledge-base relation are regarded to express that relationship if they belong to the same sentence. To avoid redundant information, we also set a series of rules to only keep salient nodes in the graph. We believe a graph built this way can capture the global structure information of the original text. We then combine a Bi-LSTM and a Graph Attention Neural Network to encode the graph. Having this vital information in mind, we design a novel decoder to take advantage of information from both the original document and the graph.
    
    \citet{sharma-etal-2019-entity} tries to use the information of entities when extracting salient sentences. However, this method is based on a two-step procedure and does not take the relation between entities into consideration. In order to mitigate the long-distance relationship problem, \citet{fernandes2019structured} uses graphs to model highly-structured data. Coreferences of entities are connected with a special REF edge, but the relationships between different coreference clusters and entities are not considered. The model still has difficulty identifying relationships between entities that belong to different coreferences.
\paragraph{Our Contributions} We propose a new architecture, which we call DSGSUM, to integrate graph information in the standard encoder-decoder framework for abstractive summarization. The contribution of our work is in three-folds:
(i) To the best of our knowledge, we are the first to leverage the relationship between different coreferences or entities in summarization tasks, which improve the ability of our model to comprehend complex relationships between entities.
(ii) We develop a new method to build graphs from text for the summarization task, which is proven to capture the global structural information better than previous methods according to our experiments.
(iii) We design a novel decoder to incorporate the graphical information, achieving new state-of-the-art results in abstractive summarization, which can serve as a stepping stone for future research.

    \begin{figure*}[t]
    \centering
    \includegraphics[height=8cm,width=15cm]{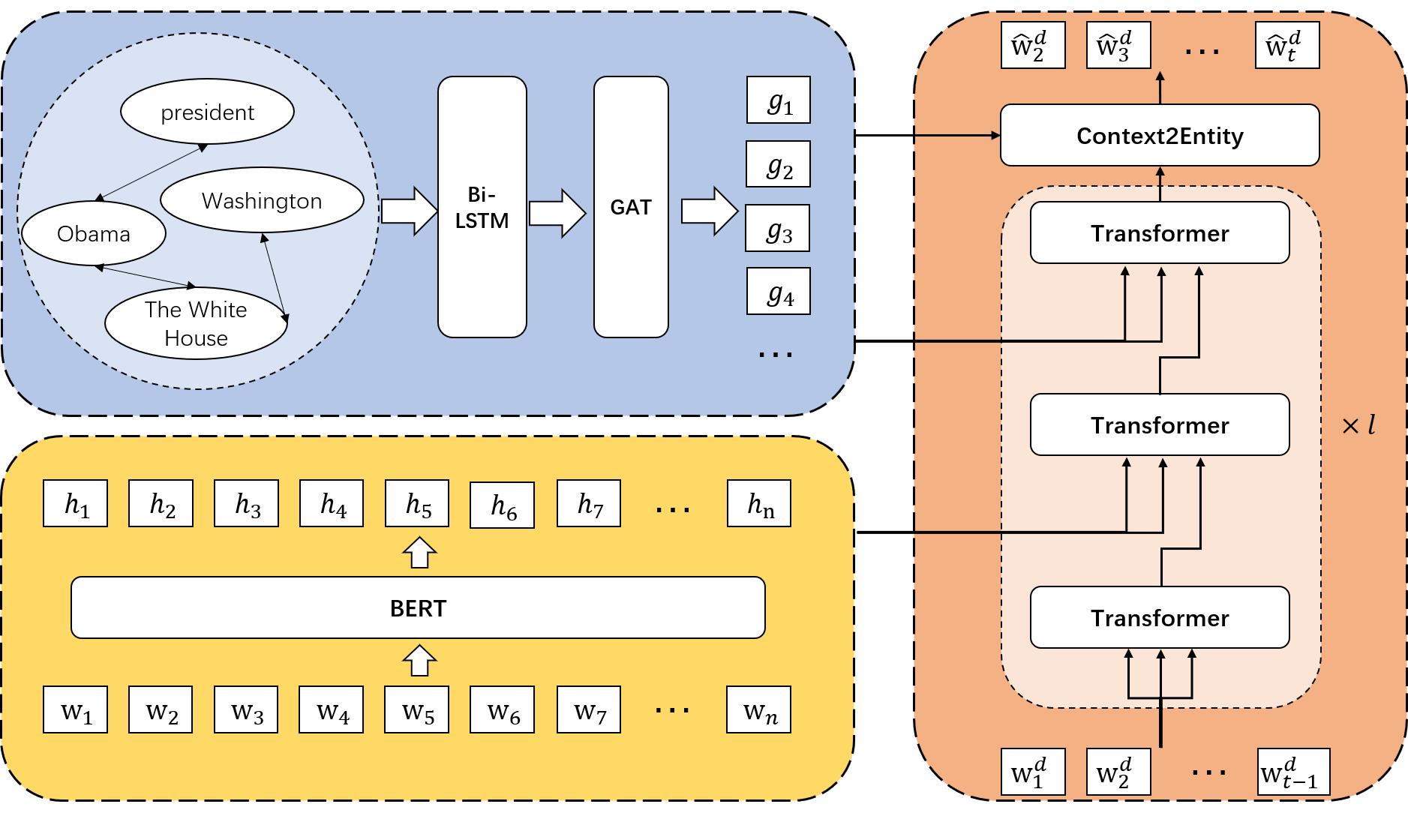}
    \caption{The framework of DSGSUM: (1) a graph encoder to obtain the contextualized representation of each graph node, (2) a BERT-based word representation encoder, and (3) a novel decoder structure which combines Transformer with Contxt2Entity attention to generate the final output $\widehat{\mathbf{w}}^d$, utilizing the structural information.}
    \label{fig:structure}
    \end{figure*}

\section{Model}
    In this section, we introduce our DSGSUM model that leverages graphic information in the abstractive summarization framework. Figure \ref{fig:structure} shows three modules of our architecture: (1) a graph encoder to obtain the contextualized representation of each graph node, (2) a BERT-based word representation encoder similar to \citet{liu-lapata-2019-text}, and (3) a novel decoder structure which incorporates the graph information.
    
    Suppose we have a document $D$ with $n$ words. The document can be represented as $D=[x_{1}, x_{2}, \cdots,x_{n}]$, where $x_i$ represents the $i$-th word in the document. Each word $x_i$ can be embedded into a $p$-dimension vector $\mathbf{w}_i$, and we denote the embedding of input tokens at the decoding process as $\mathbf{w}^d$. Note that we insert [CLS] and [SEP] at the beginning and end of each sentence respectively. A chunk of consecutive words makes up an entity, which is a node in the graph structure. For example, the $j$-th entity consists of words from  $x_a$ to $x_b$ can be denoted as $\tilde{e}_j=[x_a,\cdots,x_b]$. To obtain the representation $\mathbf{g}$ of each graph node, we firstly use a Bi-LSTM to encode each entity as $\mathbf{e}$. Then we use a graph encoder to take the relationship into consideration, which derives the final representation $\mathbf{g}$ of each node. To obtain the encoded representation $\mathbf{h}$ of each word in the document, we adopt a pre-trained BERT model. In the prediction process, we apply the normal Transformer decoding process first to obtain the hidden state $\mathbf{s}$. Then we adopt a context-to-entity attention inspired by Bi-Directional Attention Flow (BIDAF) \citep{Seo2017BidirectionalAF} to leverage the structured information.
 
    \subsection{Graph Extraction}
    Before we introduce the model structure, it's necessary to clarify the procedure to generate the graph for each document. We use existing state-of-the-art tools to infer entities and extract the relation inspired by distant supervision.
    
    \paragraph{Entity Extraction} We use Spacy \cite{spacy2}, a natural language processing tool with industrial-strength to process each document to extract entities $\tilde{e}_i$. As for coreference extraction, we use neuralcoref\footnote{\url{https://github.com/huggingface/neuralcoref}} developed by huggingface. The result is in the form of clusters $\tilde{c}_j$, consisting of two or more entities in a document referring to the same object (e.g. a person). For example, a cluster $\tilde{c}_j$ consists of two mentions $\tilde{e}_1,\tilde{e}_2$ can be represented as $\tilde{c}_j=[\tilde{e}_{j,1},\tilde{e}_{j,2}]$, and the first entity is typically called the main mention of this cluster. Although these entities can provide our system with extra-linguistic information, trivial entities can discourage our model from concentrating on salient entities which are vital to detect crucial content. In order to avoid introducing redundant information, we only keep entities that are not stopwords. Moreover, a coreference cluster will be ignored if the total number of mentions is less than 3.
    
    \paragraph{Relation Extraction}
    There are quite a few high-quality relation extraction models, but most of them can only be applied to a specific domain with annotated relation labels. However, the summarization dataset lacks golden relation annotations, therefore it's difficult to build up a model to automatically extract relations. In this paper, we mainly consider two types of relationships:\\
    (1) Distant supervision \cite{mintz-etal-2009-distant} is proposed to address the shortage of golden relationship annotations. Its idea comes from the fact that any sentence containing a pair of entities that exist in a known knowledge-base relation are likely to have that type of information. Large scale graphical knowledge databases such as ConceptNet \citep{speer-havasi-2012-representing} use graphical structure to express intricate relations between concepts. In our work, we use ConceptNet to decide whether there is a relation between two entities. However, the relation of triples extracted by this method are often redundant or trivial, so we limit the relationship types to `UsedFor', `CapableOf', `Causes', `CausesDesire', `Desires', and `ObstructedBy'. \\
    (2) Also inspired by distant supervision, we propose that two main mentions of different clusters have relations if they are in the same sentence. We notice that the object that a cluster refers to is of great importance because it appears in the original text more than twice. Additionally, the main mention of a cluster with more quotations in the source document is more likely to appear in the golden summary \citep{sharma-etal-2019-entity}. Most of the previous papers simply label different mentions in one cluster as coreference. These methods neglect the relationship between different clusters which reflects the interaction of salient mentions. We propose that if two entities come from different coreference clusters and appear in the same sentence, then the two clusters can be deemed to have a relationship. For example, `Barack Obama' and `the White House,' appear many times in the training documents, although not always in the same form. Then it is reasonable to infer that 'Barack Obama' and 'the white house' have a strong relationship if they are quoted in the same sentence. 
    
    Combining the entity extraction and the relation extraction, we are able to build a graph representation for each document. 
    
    \subsection{Graph Encoder}
    Next, we use an LSTM \cite{10.1162/neco.1997.9.8.1735} to encode entities and apply a Graph Attention Network to capture relationship information of the graph we get from last step.
    
    \paragraph{LSTM Encoder}
    Entity span $\tilde{e}_i$ is often multiple words as shown in the Entity Extraction section. The initial vector representation $\mathbf{e}$ should be obtained before applying the graph structure model. The representations of the start word and the end word of a span typically contain most of these mention’s information \citep{lee-etal-2017-end}. However, our experiment shows that simply using the last word works better here. We run a Bi-LSTM over the initial embedding of each word in entity span $\tilde{e}_i$, and the concatenation of backward and forward outputs of the last word is defined as $\mathbf{e}_i$.

    \paragraph{Graph Attention Network}
    The structure of our graph encoder is similar to the Graph Attention Network (GAT) which enables efficient and parallel computation. The hidden representation of each node in a graph is obtained by attending over its neighbors based on the Transformer structure \citet{NIPS2017_3f5ee243}. In the first iteration, nodes in the GAT can only interact with their nearest neighbor nodes. We run the GAT more than twice to obtain more global representation for each node.
    
    In the $l$-th iteration, the representation $\mathbf{g}^l_i$ of $\tilde{e}_i$ is contextualized by attending over its neighbor nodes. Before the model goes on with the next iteration, the derived vector is processed through a specially designed attention mechanism:
    \begin{equation}\label{}
    \mathbf{g}^l_i= \mathtt{FFN}(\mathtt{LayerNorm}(\mathbf{g}_i^{l-1}+ \sum_{j \in \mathcal{N}_{i}} \alpha_{i j} \mathbf{g}_{j}^{l-1}))
    \text{,}
    \end{equation}
    \begin{equation}\label{}
    \alpha_{i j} = \frac{\mathtt{exp}((W_1 \mathbf{g}_i) (W_2 \mathbf{g}_j)^{T})} {\sum_{k \in {\mathcal{N}_i }} \mathtt{exp}((W_1 \mathbf{g}_i) (W_2 \mathbf{g}_k)^{T})} \text{.}
    \end{equation}
    Where $ \mathcal{N}_i$ denotes the neighborhood of $\tilde{e}_i$, including $\tilde{e}_i$ itself; $W_1$ and $W_{2}$ denote trainable matrices parameters throughout the paper. $\mathrm{FFN}$ is a two-layer feedforward network and $\mathrm{LayerNorm}$ is a layer normalization operation.
        
    In this paper, we adopt an N-headed self-attention setup, with N independent attentions calculated and concatenated before a residual connection is applied. The final vector representation of $i$-th graph node is denoted as $\mathbf{g}_i$. \citet{tay2020synthesizer} propose that random alignment attention matrices surprisingly perform quite competitively. We are interesting to apply this trick in our method, but the results are not improved. Therefore, we simply adopt the vanilla transformer in the paper.
    
    \paragraph{Frequency Embedding} We notice that the entities that appear frequently in the original text are more likely to appear in the golden summarization, which means that these entities are more important in the process of generating a summary. In order to better detect these salient entities, we additionally introduce a frequency embedding for these derived entity representations, which is similar to the idea of positional embeddings in BERT. We sort entities in descending order according to the number of occurrences in the original text. Before being fed into the graph encoder, the representation of each entity is defined as $\mathbf{g}_i^0:=\mathbf{e}_i+\mathrm{FE}(i)$, where $\mathrm{FE}(\cdot)$ denotes the pre-initialized frequency embedding, and $i$ represents the ranking of the entities after sorting.
    
    \subsection{Word Representation Encoder}
    Motivated by \citet{liu-lapata-2019-text}, we use BERT to encode the document. Each token $w_i$ is assigned three kinds of embeddings: a token embedding indicating the meaning of each token, a segmentation embedding discriminating between two sentences, and a position embedding denoting the position of each token. The encoder structure in this paper is similar to BERTSUMABS \citep{liu-lapata-2019-text}. The encoded embedding of each word is denoted as $\mathbf{h}=\mathrm{Transformer}(\textbf{w}, \textbf{w}, \textbf{w})$. For simplicity, the first place denotes query vector, and the second and third place denote the key and value vector respectively.

    \subsection{Decoder}
    In a traditional seq-to-seq Transformer structure, the decoder obtains its contextualized representation $\textbf{s}^d$ in a way similar to the encoder,  with a modification to prevent the model from attending to subsequent positions. There is also a sub-layer to help the decoder focus on appropriate places in the input sequence. More details can be found in \citep{NIPS2017_3f5ee243}. Here we simply denote the output of this decoder as:
    \begin{equation}\label{}
    \textbf{s}^d=\mathrm{Transformer}(\textbf{w}^d, \textbf{w}^d, \textbf{w}^d)
    \text{,}
    \end{equation}
    \begin{equation}\label{}
    \textbf{s}=\mathrm{Transformer}(\textbf{s}^d, \textbf{h}, \textbf{h})
    \text{.}
    \end{equation}
    
    \paragraph{Context-to-Entity Attention}
    The decoder state $\mathbf{s}_t$ is supposed to be entity-aware. In machine comprehension, \citet{Seo2017BidirectionalAF} introduce the
    Bi-Directional Attention Flow (BIDAF) network, a multi-stage hierarchical process that represents the context at different levels of granularity and uses a bidirectional attention flow mechanism to obtain a query-aware context representation. Inspired by this work, we propose a \textbf{context-to-entity} attention in this paper. In order to avoid redundant information our system, we try to introduce entity filter module before context2entity. However, the performance is below expectation, so we left our attention mechanism to decide which entity should be paid more attention.
    
    In this module, we compute attentions from context to entities, which is derived from a similarity matrix, $\mathbf{M}:=(M_{ti})_{1\le t\le T, 1\le i\le I} \in \mathbf{R}^{T \times I}$, between the contextual embeddings of the decoder word $\mathbf{s}$ and the entity $\mathbf{g}$, where $T$ and $I$ are size of $\mathbf{s}$ and $\mathbf{g}$, respectively. Each element $M_{ti}$  indicates the similarity between the $t$-th decoder hidden vector and the $i$-th graph node, which is calculated by:
    \begin{equation}\label{}
    M_{t i}=\alpha\left(\mathbf{s}_{t}, \mathbf{g}_{i}\right)
    \text{.}
    \end{equation}
    where $\mathbf{s}_{t}$ is the $t$-th vector in decoder state, $\mathbf{g}_{i}$ is the embedding of the $j$-th entity. $\alpha$ is a function defined as $\alpha(\mathbf{s},\mathbf{g})=W_s\mathbf{s}+W_g\mathbf{g}+W_{sg}(\mathbf{s}\circ \mathbf{g})$, and $\circ$ denotes element-wise multiplication. Context-to-entity attention signifies which entities are most relevant to decode state $\mathbf{s}$ respectively. The attention weight is computed by , and the weighted entity vector is $\tilde{\mathbf{g}}_{t}=\sum_{i=1}^{I} a_{ti}\mathbf{g}_i$, which is a $d$-dimensional vector. The final graph-enhanced representation $\mathbf{s}^g$ can be defined as:
    \begin{equation}\label{}
    \mathbf{s}^g=W_G[\mathbf{s};\tilde{\mathbf{g}};\mathbf{s}\circ \tilde{\mathbf{g}}]
    \text{,}
    \end{equation}
    With the decoder state embedding $\mathbf{s}^g$ at hand, we can simply predict the word distribution $P_{vocab}$ as:
    \begin{equation}\label{}
    P_{vocab}=\mathrm{softmax}(W_P\mathbf{s}^g)
    \text{.}
    \end{equation}
    where $W_P$ projects the hidden state to the size of vocabulary. Here we want to mention that we also tried to add a entity-to-context attention flow to make a complete BIDAF, while the performance actually got worse. Experiments show that a single context-to-entity attention is good enough for our system. We also tried to introduce copy mechanism to copy salient entities to the generated summary. However, through ablation study, we found that it actually hurt the performance. This might be because the encoder based on powerful pre-trained model BERT has already learned part of the salient entities information, and copying redundant entities will add noise to the learning process. More details of this part can be found in the Appendix.
    
    \section{Experimental Setups}
    \paragraph{Datasets and Preprocessing} We perform our experiments on the non-anonymous CNN/Dai-lymail (CNNDM) dataset\footnote{\url{https://cs.nyu.edu/~kcho/DMQA/}} ~\citep{nallapati-etal-2016-abstractive}. We also adopt the New York Times dataset ~\citep{durrett-etal-2016-learning}, which licensed by LDC\footnote{\url{https://catalog.ldc.upenn.edu/LDC2008T19}}.  The source documents in the CNNDM training set on average have 766 words spanning 29.74 sentences while for summaries the numbers are 53 words and 3.72 sentences. We used the non-anonymized version of the CNNDM corpus and truncated source documents to 512 tokens. We followed the preprocessing steps in ~\citep{nallapati-etal-2016-abstractive}, obtaining 287,227 training pairs, 13,368 validation pairs and 11,490 test pairs.
    
    For New York Times (NYT), we adopted the same split strategy as ~\citet{durrett-etal-2016-learning}, and partitioned the whole dataset to 100,834 training pairs, 9,706 validation pairs, and 4000 test pairs. The test set contains all articles published since January 1, 2007. Then following ~\citet{durrett-etal-2016-learning}, we created the NYT50 dataset by removing the pairs with summaries shorter than 50 words, and obtained a filtered test set with 3,452 examples. Input documents were truncated to 800 tokens, and all documents were tokenized with the Stanford CoreNLP toolkit \citep{manning-EtAl:2014:P14-5}.

    \paragraph{Training Details and Parameters.}
    Both source and target texts were tokenized with BERT's subword tokenizer.  We rarely observe issues with out-of-vocabulary words in the output thanks to the subword tokenizer. We used Spacy \citep{spacy2} to extract entities using model version `en-core-web-sm'. As for coreference extraction, we used neuralcoref developed by huggingface. 
    
    After getting the graph, we first used a 2 layer Bi-LSTM with 384 hidden nodes to encode the entities, with the dropout rate to be 0.2. For the graph encoder, we adopted a 4-headed transformer implementation, where the number of graph layers is 3.  We applied dropout with probability 0.2 before all linear layers. We adopted the transformer model implementation by \citep{Wolf2019HuggingFacesTS}, and BERT-based models were used throughout the experiments. Our Transformer decoder has 768 hidden units and the hidden size for all feed-forward layers is 2,048. During decoding, we used beam search with size 5 and decoded until an end-of-sequence token is emitted. In order to reduce repetitions, repeated trigrams were blocked \citep{paulus2018a}. We used two Adam optimizers for the encoder and the decoder, respectively, each with the same warmup-steps and learning rate settings as \citet{liu-lapata-2019-text}. The pre-trained encoder was ﬁne-tuned with a smaller learning rate and smoother decay.
    
    We chose the top-3 checkpoints based on their validation loss and reported the averaged results of these checkpoints on the test set. All of our models were trained for 150,000 steps on 2 GPUs (Tesla V100) with gradient accumulation every two steps, which roughly takes about a day and a half. We reported Rouge-1, Rouge-2, and Rouge-L as evaluation metrics of our models. Rouge-1 and Rouge-2 measure the unigram and bigram overlap respectively. Rouge-L is the longest common subsequence for assessing fluency~\citep{durrett-etal-2016-learning}.

    \begin{table}[t]\scriptsize
       \centering
               \caption{ROUGE F1 results of various models on CNNDM test set using full-length F1 Rouge-1 (R1), Rouge-2 (R2), and Rouge-L (RL). For NYT50, we used limited-length ROUGE Recall \cite{liu-lapata-2019-text}, where predicted summaries are truncated to the length of the gold summaries. We applied bootstrap resampling technique \citep{berg-kirkpatrick-etal-2012-empirical}, and our model performs significantly better than other models with $p<0.05$.}
        \begin{tabular}{m{1cm}<{\centering}m{6.0cm}m{1cm}<{\centering}m{1cm}<{\centering}m{1cm}<{\centering}}
            \toprule
            Dataset & Model          & R1    & R2    & RL    \\ \midrule
            \multirow{9}{*}[-2pt]{CNNDM}
            & LEAD-3         & 40.42 & 17.62 & 36.67 \\
            & POINTGEN+COV\citep{see-etal-2017-get} & 39.53 & 17.28 & 36.38 \\
            & REWRITE\citep{chen-bansal-2018-fast} & 40.88 & 17.80 & 38.54 \\
            & DEEPREINFORCE\citep{paulus2018a} & 41.16 & 15.75 & 39.08 \\
            & BOTTOMUP\citep{gehrmann-etal-2018-bottom} & 41.22 & 18.68 & 38.34 \\
            & SENECA\citep{sharma-etal-2019-entity} & 41.52 & 18.36 & 38.09 \\
            & DCA\citep{celikyilmaz-etal-2018-deep} & 41.69 & \textbf{19.47} & 37.92 \\
            & BERTSUMABS\citep{liu-lapata-2019-text} & 41.72 & 19.39 & 38.76 \\
            & \textbf{DSGSUM} & \textbf{41.96} & {19.29} & \textbf{38.98} \\
            \midrule
            \multirow{6}{*}[-2pt]{NYT50}
            & LEAD-3         & 39.58 & 20.11 & 35.78 \\
            & POINTGEN\citep{see-etal-2017-get} & 42.47 & 25.61 & --- \\
            & POINTGEN+COV\citep{see-etal-2017-get} & 43.71 & 26.40 & --- \\
            & DEEPREINFORCE\citep{paulus2018a} & 42.94 & 26.02 & --- \\
            
            & BERTSUMABS\citep{liu-lapata-2019-text} & 48.92 & 30.84 & 45.41 \\
            & \textbf{DSGSUM} & \textbf{49.86} & \textbf{31.04} & \textbf{46.64} \\
            \bottomrule
        \end{tabular}

        \label{table:nyt}
    \end{table}

    \paragraph{Baselines and Comparisons.}
    Besides baseline LEAD-3, we compared our model with existing popular state-of-the-art abstractive summarization
    models on CNNDM datasets: (1) pointer-generator model with coverage (POINTGEN+COV) \citep{see-etal-2017-get}; (2) sentence rewriting model (REWRITE) \citep{chen-bansal-2018-fast}; (3) RL-based abstractive summarization (DEEPREINFORCE) \citep{paulus2018a}; (4) bottom-up abstraction (BOTTOMUP) \citep{gehrmann-etal-2018-bottom};
    deep communicating agents-based summarization (DCA) \citep{celikyilmaz-etal-2018-deep}; a system for Entity-driven coherent abstractive summarization (SENECA) \citep{sharma-etal-2019-entity}, and an extended version of BERT for summarization (BERTSUMABS) \citep{liu-lapata-2019-text}. For NYT50, we replicate the socore of POINTGEN and DEEPREINFORCE from \citep{sharma-etal-2019-entity}.

    \section{Results}
    \subsection{Automatic Summary Evaluation}
    Table \ref{table:nyt} summarizes the performances of all abstractive models on both CNN and NYT50 dataset. Our DSGSUM outperforms other methods by a large margin. Specifically, we can observe that DSGSUM beats the previous state-of-the-art model BERTSUMABS on R-1 and R-L. To verify the performance gap between DSGSUM and BERTSUMABS, we run another 4 training runs. We observe that the variance of the different models is non-overlapping, and all runs of DSGSUM outperformed BERTSUMABS on metrics of R-1 and R-L. It has been shown that the pretrained language model could capture a surprising amount of world knowledge. BERT may already have the information about the entities and the relations between them. However, the knowledge is stored implicitly in the parameters of a neural network \citep{guu2020realm}. The result shows that the design of our model that extracts the salient entities and models their relations explicitly does bring extraction gains for the generation system. If the information is trivial, incorporation of the information will not bring significant improvement, and even spoil the model performance.

    Under R-2 metric, though not signiﬁcant, the score of our model is slightly lower than BERTSUMABS on CNNDM. This may partly due to the fact that our model focuses more on the logical relationships of the full text, causing the model to ignore some local grammars. As we will discuss in section \ref{sub:human}, BERTSUMABS also performs better on grammar on cnndm. Grammar often determines what kind of words should appear after the current word. There may be a positive correlation between R-2 score and grammatical performance. Consequently, BERTSUMABS performs better on R-2 score, which measures bi-gram overlap.

    \begin{figure*}[t]
    \centering
    \includegraphics[height=5cm,width=16cm]{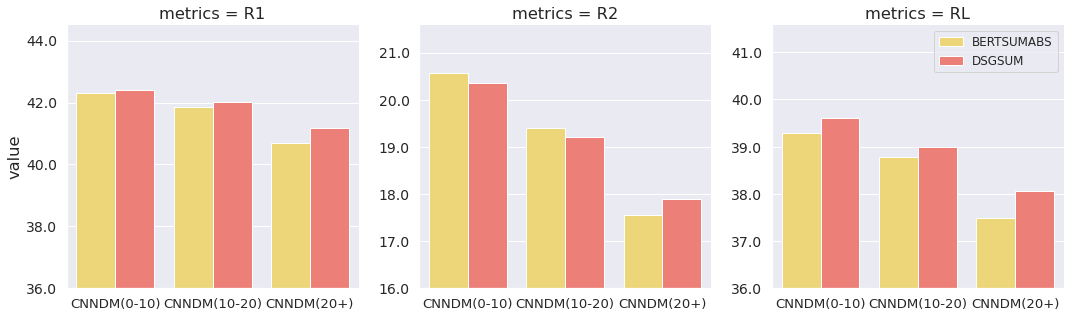}
    \caption{Results of our models and BERTSUMABS on CNNDM test set by the number of entities, using full-length F1 Rouge-1 (R-1), Rouge-2 (R-2), and Rouge-L (R-L). The split dataset has 3622, 5338, 2530 samples respectively. Our model is better than BERTSUMABS on all split datasets, and has the most significant improvement in the divided dataset containing more entities (e.g. CNNDM(20+)).}
    \label{fig:compare}
    \end{figure*}
    
    \subsection{Ablation Study}
     The extracted mentions are of great importance, however the relationship between them can not be ignored. We believe that the writer would not mention two salient subjects in the same sentence if they do not have a strong correlation. For example, the coreference clusters `Barack Obama' and `the White House' are mentioned at least twice at different sentences and locations in the document, so the relationship between their main mentions can be deemed to have a long-range dependency, which is helpful to generate more coherent summaries \citep{sharma-etal-2019-entity}.
    \begin{table}[t]\scriptsize
           \centering
            \caption{Results of ablation study on CNNDM. "- gat" denotes not to use graph attention network, in which the entities are only encoded by the lstm encoder; "- context2entity" denotes not to use the entity information while generating summary, and the results is almost the same as BERTSUMABS ; "- FE" denotes not to use frequency embedding, in which the frequency embedding is not added into the representation of each entity.}
        \begin{tabular}{p{1.0cm}p{5.0cm}p{0.9cm}<{\centering}p{0.9cm}<{\centering}p{0.9cm}<{\centering}}
            \toprule
            Dataset& Model          & R1    & R2    & RL    \\ \midrule

            \multirow{4}{*}[-2pt]{CNNDM} &{DSGSUM} & \textbf{41.96} & \textbf{19.29} & \textbf{38.98} \\
            &{DSGSUM - gat} & {41.83} & {19.19} & 38.81 \\
            &{DSGSUM - context2entity} & 41.74 & 19.15 & 38.79 \\
            &{DSGSUM - FE} & 41.77 & 19.14 & 38.79 \\
            \bottomrule
        \end{tabular}

        \label{table:ablation}
    \end{table}
    To further highlight the value of using graphs rather than only salient entities, we propose an entity-version of our model by dropping the graph attention network. In this model, the embedding of each entity won't reflect the information of their relation. Both models make use of the same set of entities for the same test sample. The score on CNNDM is still better than BERTSUMABS. As shown in Table \ref{table:ablation}, due to the lack of information on relations, the performance is down to 41.83/19.19/38.81. We can conclude that related entities improve summary generation compared to an unrelated collection of entities. The result indicates that the introduction of the relationship between nodes is of great benefit for the model to detect salient information and transform this information into the surface form in the correct way. In addition, we conducted comparative experiments showing that  context2entity and frequency embeddings are also effective.

    \begin{table*}[t]
    \centering
        \caption{Sample golden summary, extracted entities and summaries generated by BERTSUMABS and DSGSUM. Different colored fonts indicate overlap with corresponding extracted entities. The results show that most extracted entities appear in the golden summary and also exist in the summary generated by DSGSUM. It verifies that entities are important for generating summaries and our model could properly use these entities while generating summaries.}
    \small
    \begin{tabular}{|l|}
    \hline
    \begin{tabular}[c]{@{}l@{}}\textbf{Entity}:\\ \textcolor{blue}{boston}, \textcolor{brown}{the plane}, \textcolor[rgb]{0.5,0.0,0.0}{delta},  \textcolor{green}{new jersey}, \textcolor{purple}{newark}, \textcolor{yellow}{flight 271 from paris}, \textcolor{red}{logan international in boston},\\  new york, \textcolor{orange}{king kong}, another female passenger, \textcolor[rgb]{0.0,0.0,0.5}{an emergency landing in boston}, one passenger,\\  \textcolor[rgb]{0.6,0.4,0.8}{the turbulence}\end{tabular}                                                                                                                           \\ \hline
    \begin{tabular}[c]{@{}l@{}}\textbf{Gold}:\\ \textcolor[rgb]{0.5,0.0,0.0}{delta} said `a small number of customers ' on \textcolor{yellow}{flight 271 from paris} to \textcolor{purple}{newark} complained of nausea \\ and possible minor injuries. two people taken to massachusetts general hospital after  \textcolor[rgb]{0.0,0.0,0.5}{unscheduled} \\  \textcolor[rgb]{0.0,0.0,0.5}{landing} at \textcolor{red}{logan airport in boston}. extreme \textcolor[rgb]{0.6,0.4,0.8}{turbulence} prevented \textcolor{brown}{the plane} from landing at \textcolor{purple}{newark} \\ and then jfk before pilot headed to \textcolor{red}{logan}.\end{tabular} \\ \hline
    \begin{tabular}[c]{@{}l@{}}\textbf{BERTSUMABS}:\\ \textcolor[rgb]{0.5,0.0,0.0}{delta} \textcolor{yellow}{flight from paris} , france to \textcolor{purple}{newark} , \textcolor{green}{new jersey} was diverted to \textcolor{blue}{boston}. two people were \\ taken to massachusetts general  hospital with what are believed to be minor injuries. one passenger \\ said  it felt like `\textcolor{orange}{king kong} picked up \textcolor{brown}{the plane}  and shook  it like there was no tomorrow'\end{tabular}                                              \\ \hline
    \begin{tabular}[c]{@{}l@{}}\textbf{DSGSUM}:\\  \textcolor[rgb]{0.5,0.0,0.0}{delta} said `a small number of customers ' on \textcolor{yellow}{flight 271 from paris} to \textcolor{purple}{newark} airport in  \textcolor{green}{new jersey} \\ complained of nausea  and possible minor injuries. two people were taken to massachusetts general \\ hospital with what are believed to be minor injuries after the  \textcolor[rgb]{0.0,0.0,0.5}{unscheduled landing} at \textcolor{red}{logan airport} \\ \textcolor{red}{in boston}.\end{tabular}                      \\ \hline
    \end{tabular}

    \label{table:case}
    \end{table*}

    \begin{table}[t]\scriptsize
           \centering
                \caption{Results of human evaluation on CNNDM. Best results are in bold.}
        \begin{tabular}{m{2cm}m{2cm}<{\centering}m{2cm}<{\centering}m{2cm}<{\centering}}
            \toprule
             & RELEVANCE & GRAMMAR & COHERENCE   \\ \midrule
            HUMAN & 4.83 & 4.90 & 4.83 \\
            \cdashline{1-4}
            BERTSUMABS & 4.33 & \textbf{4.86} & 4.07 \\
            DSGSUM & \textbf{4.57} & {4.63} & \textbf{4.43} \\
            \bottomrule
        \end{tabular}

        \label{table:human}
    \end{table}
    \subsection{The Value of Structured Information}
    We collect the entities in the summary written by humans. The generated summary of our model can cover about 62.01\% of these entities on average, which is higher than that of BERTSUMABS (60.11\%). Entities that appear in the golden summary are salient, often revealing the main topics of a sentence. The results show that our model can identify these important mentions and present them in the generated summary properly.

    \paragraph{Why to Model Structured Data Explicitly}
    We hold the idea that a system provided with structured information explicitly performs better. To verify this hypothesis, we split the test set according to the number of the extracted entities from the initial document and compare our model with BERTSUMABS.  The entities we extract mainly include mentions from coreferences and entities recognized by Spacy. If a document contains many coreferences, it means that the logical relations in this article is more complicated. Most existing models are poor at comprehending such data sets and generating summaries. Provided explicitly with this information, it is expected that the benefit of our models will be more evident for those samples containing more entities. The result in Figure \ref{fig:compare} shows that both our model and BERTSUMABS perform best on the divided dataset which has entities lower than 10. There are fewer entities in these kinds of documents, and models easily identify the main objects of these articles. When the entities extracted from the article exceed 20, both models perform the worst. But because our model explicitly utilizes structured information, our model performs significantly better than BERTSUMABS in this type of article.

    \subsection{Human Evaluation}\label{sub:human}
    Firstly, human evaluation is conducted to analyze the informativeness and readability of the summaries
    generated by our models. We conducted a manual evaluation with 3 proficient English speakers. We primarily used the following criteria to assess the generated summaries: \textbf{relevance}-whether the generated summary has strong relevance with the golden summary; \textbf{grammar}, \textbf{coherence}-whether the summary presents contents and entities in a coherent manner. To conduct the evaluation, we randomly selected 30 examples from the CNNDM testing set and asked the volunteers to subjectively rank the summaries against each other on a scale of 1 (worst) to 5 (best). For simplicity, we only provide the extracted entities to assist volunteers to judge, not the whole set of triples. Each example consisted of an article and three summaries, i.e., the golden summary, BERTSUMABS, and our proposed DSGSUM. Obviously, the summaries were randomly shuffled, and the model used to produce each summary was unknown to prevent bias. The results are presented in Table \ref{table:human}. Our model performs well on relevance and coherence. Surprisingly, BERTSUMABS ranks higher on GRAMMAR when compared to DSGSUM. Through manual inspection, we find that the most common cause of grammatical errors is the lack of adjacent components of words in a sentence, which can also lead to the lower bi-gram overlap compared with BERTSUMABS. A sample of case study is shown in Table \ref{table:case}.

    \section{Conclusion}
    In this paper, we propose a new architecture, which we call DSGSUM, to integrate structured information for abstractive summarization. We develop a new method to build graphs for the summarization task and design a novel decoder to leverage the graphical information.
    Our model performs better than previous state-of-the-art models on the CNNDM and NYT50 datasets.
    We have shown that structured information is advantageous for abstractive summarization models.
    In this paper, our methods to extract triples also introduce noises and we have formulated some rules to filter triples. With more advanced methods to filter those noises, our model may perform better. And copy mechanism may also be helpful for our model, which directly copies important entities from the article into the abstract.  We leave these parts to future work. More details can be found in the appendix.

    \section*{Supplementary Material}
    Supplementary material online include: data link, python code and an instruction file needed to reproduce the results; an appendix containing additional structures and experiments we have tried. The web link is https://github.com/martin6336/DSGSum.

\bibliographystyle{jds}
\bibliography{jds2020.bib}

\end{document}


\appendix
    \section{Supplementary}

    \paragraph{Token-independent attention}
    In the vanilla self-attention model, the attention weight is based on the pairwise token interactions. We try a latest trick, in which the attention weights do not depend on any input tokens. Instead, the attention weights are initialized to random values. These values can then either be trainable or kept fixed \citep{tay2020synthesizer}. We combine this token-independent weight with the vanilla weight matrix in Transformer and use the following attention calculation equation instead:
    \begin{equation}
    \mathrm{Attention}(Q,K,V)=(\mathrm{softmax}(\frac{Q K^{T}}{\sqrt{d_{k}}})+G)V.
    \end{equation}
    where $G$ is a random initialized global weight matrix and trained with the model. We implement this idea in the graph attention network, and the ROUGE F score is 41.90/19.27/38.94.
    
    \paragraph{Entity filter module}
    In order to avoid redundant information in our system, we try to introduce the entity filter module before Context2Entity. We tried two method. In the first method, we simply apply the dense connection to the entity vector to get the salient score for each entity. In the second method, we apply attention mechanism between entities and decode state $\mathbf{s}_t$, this derive $\mathrm{score}_{i,t}=\mathbf{g}_i^T\mathbf{s}_t$. Therefore we get the salient score $\mathrm{salience}_i=\sum_t{\mathbf{s}_t}$ for each entity.  Then select the top-k entities to go on with Context2Entity module. However, the performance is below expectations, so we apply the attention mechanism to let the model decide which entity should be paid more attention.
    
    \paragraph{Usage of graph information}
    There is another way to utilize the graph information in the prediction process. We use the hidden state as query, the graph node representation as key and value. This derive the final graph-enhanced representation $\mathbf{s}^g$:
    \begin{equation}\label{}
    \textbf{s}^{g}=\mathrm{Transformer}(\textbf{s}, \textbf{g}, \textbf{g}).
    \end{equation}
    But this method performs slightly worse than our Context2Entity attention.
    
    \paragraph{Entity-to-context attention}
      Entity-to-context attention aims to detect which decoder state has the closest similarity with each entity. It should be noticed that current state is unaware of subsequent states, thus we modify the initial BIDAF by masking the followed state for each state. We obtain the attention weights by $b_t=softmax({max}_{col}(\mathbf{M}_{:t,:}))$, the weighted state vector is $\tilde{\mathbf{s}}_{t}=\sum_{i=1}^{t} b_{i}\mathbf{s}_i$.
    
    \paragraph{Copy mechanism}
    In the prediction process, we also try to apply copy salient words or entities from the input document but the results are not well. We infer that the input of the summarization task is often long documents, implementing copy mechanism directly may not get satisfying results. \citet{gehrmann-etal-2018-bottom} use mask and modify the copy attention distribution to only include tokens identified by the content selector. Inspired by this idea, we join the tokenized entities with the [SEP] token, which produces an entity text $C$ similar to the input document $D$. Then we use the document encoder to get embedding $c_i$ of every token in this text.
    
    For a given token representation $c_i$ in $t$-th decode process, we concatenate $c_i$ with current decoder state $\mathbf{s}^g_t$, and calculate the similarity score of these two vector:
    \begin{equation}\label{}
    \beta_t^i=W_c[\mathbf{s}^g_t;c_i].
    \end{equation}
    We adopt the basic pointer generator structure which combines two probabilities, one comes from the pointer network $P_{copy}(c_i)=\frac{e^{\beta_t^i}}{\sum_j{e^{\beta_t^j}}}$ and the other is the normal generation probability $P_{vocab}$. Then a switch probability $p_{gen}=\sigma(w_g \mathbf{s}^g_t +w_c \mathbf{c}+b_{gen})$ combine these two distribution and determine the final probabilities of each word in vocabulary. But the results is still not up to our expectations. We left this part to the future work.

    \paragraph{Significant test} The size of the CNNDM test set is 11,490. We sample 3000 samples from the test set with replacement, and this process is repeated 100 times. And We draw 1000 samples from the nyt50 test set which has 3,452 examples. We mention in the table 1 of our paper that we adopt the method proposed in the paper "An Empirical Investigation of Statistical Significance in NLP". And the calculated p value is 0. In other words, our model performs significantly better than other models with p \textless 0.05.
    We admit that the improved score is not too much, compared with BERTSUMABS. But it is worth noting that \cite{berg-kirkpatrick-etal-2012-empirical} also propose that systems whose outputs are highly correlated will achieve higher confidence at lower metric gains.

    \paragraph{Triple filtering}The extracted entities include not only the noun phrase but some structures that modifies the major part such as articles and preposition structure. The modification parts are noisy sometimes.
    Some restricted parts are too lengthy and interfere with the model's understanding of the main part. We use a simple trick to alleviate this problem, in which we remove any triple whose argument (subject or object) has more than 10 words. The rouge score is 41.85/19.15/38.92.
The decrease in score may be that some important triples are dropped when the entity length is limited to 10. We will continue to explore the effect of the entity length limitation in the future.

\paragraph{Manual evaluation of extracted triples}
It is worth noting that this article replaces all relations with "have relation with" in order to simplify the model.
We show the extracted triples and golden summary for reference in Table \ref{table:manul}:
\begin{table}[h!]\scriptsize
      \centering
        \caption{Manual evaluation of extracted triples, with golden summarization as reference.} 
    \begin{tabular}{p{0.9\columnwidth}}
        \toprule
        extrated triples  \\ \midrule
        ('david hawkins', 'have relation with', "the organization 's commission on the use of standardized tests")
('david hawkins', 'have relation with', 'baylor university')
('david hawkins', 'have relation with', 'america')
('baylor university', 'have relation with', 'america')
('baylor university', 'have relation with', 'the sat')
('baylor university', 'have relation with', 'enough students')
('the sat', 'have relation with', 'enough students')
('act', 'have relation with', 'the sat and act scores')
('act', 'have relation with', 'act scores')
('the sat and act scores', 'have relation with', 'act scores')
('u.s. news', 'have relation with', 'u.s. news and world report')
('u.s. news', 'have relation with', 'world report')
('u.s. news', 'have relation with', 'standardized tests play an integral role in the college admissions process')
('u.s. news and world report', 'have relation with', 'world report')
('u.s. news and world report', 'have relation with', 'standardized tests play an integral role in the college admissions process')
('world report', 'have relation with', 'standardized tests play an integral role in the college admissions process')
('the rankings', 'have relation with', 'which colleges') \\ \midrule

golden summary \\\midrule

 david hawkins : admission tests are wrongly used to rank college quality. hawkins says baylor university 's incentives for test scores are a mistake. grades are much more important than test scores in admissions decisions , he says. hawkins : u.s. news should drop sat and act scores in rankings.
 \\ \midrule
 
extrated triples \\\midrule
('martin robinson a home guard', 'have relation with', 'britain')
('martin robinson a home guard', 'have relation with', 'the government')
('martin robinson a home guard', 'have relation with', 'cabinet office minister francis maude')
('britain', 'have relation with', 'the government')
('britain', 'have relation with', 'cabinet office minister francis maude')
('the government', 'have relation with', 'cabinet office minister francis maude')
('security services', 'have relation with', 'nick clegg')
('mrs may', 'have relation with', 'lib dem leader nick clegg')
('mrs may', 'have relation with', 'communications data bill')
('mrs may', 'have relation with', 'victims of serious crime , terrorism and child sex offences in the eye')
('lib dem leader nick clegg', 'have relation with', 'communications data bill')
('communications data bill', 'have relation with', 'victims of serious crime , terrorism and child sex offences in the eye')
('this bill', 'have relation with', 'victims of crime , police and the public') \\ \midrule

golden summary \\\midrule

home guard of cyber experts will protect uk business and armed forces. 9 out of 10 companies suffered online attack in 2011 - costing £ 250,000 a time.warning that cyber terrorists are targeting utility firms to disrupt supplies of gas , electricity and water. home secretary says opponents of her communications data bill must look victims of terrorism ' in the eye '. deputy pm nick clegg wants ' snooper 's charter ' delayed until 2014.\\ \midrule

extrated triples \\\midrule
('high school kindergarten mr smyth', 'have relation with', 'parents')
('high school kindergarten mr smyth', 'have relation with', 'school')
('parents', 'have relation with', 'school')
('mr smyth', 'have relation with', 'most people')
('mr smyth', 'have relation with', 'parents of new kindergarten pupils')
('mr smyth', 'have relation with', 'high school')
('mr smyth', 'have relation with', 'parents should never underestimate the old fashioned note in a lunch-box')
('mr smyth', 'have relation with', 'parents of new students')
('mr smyth', 'have relation with', 'their child')
('parents of new kindergarten pupils', 'have relation with', 'high school')
('parents of new kindergarten pupils', 'have relation with', 'parents should never underestimate the old fashioned note in a lunch-box')
('high school', 'have relation with', 'parents should never underestimate the old fashioned note in a lunch-box')
('high school', 'have relation with', 'parents of new students')
('high school', 'have relation with', 'their child')
('parents of new students', 'have relation with', 'their child')
('their child', 'have relation with', 'the teacher')
\\ \midrule

golden summary \\\midrule
teacher and education expert ciaran smyth , has revealed his top tips for making the transition from holidays to education more bearable. for high school students , he recommends creating a routine , picking a perfect study space and the occasional positive note in the lunch box. he warned parents of new kindergarten pupils to make their goodbyes quick and unemotional and label everything.
\\ \bottomrule
    \end{tabular}
    \label{table:manul}
\end{table}
\bibliographystyle{jds}
\bibliography{jds2020.bib}